# Adversarial Networks for the Detection of Aggressive Prostate Cancer

Simon Kohl[1], David Bonekamp[2], Heinz-Peter Schlemmer[2], Kaneschka Yaqubi[2], Markus Hohenfellner[3], Boris Hadaschik[4], Jan-Philipp Radtke[24], and Klaus Maier-Hein[1]

`simon.kohl@dkfz.de`

[1] Medical Image Computing, German Cancer Research Center (DKFZ), Heidelberg, Germany,
[2] Department of Radiology, DKFZ, Heidelberg, Germany
[3] Department of Medical Physics, DKFZ, Heidelberg, Germany
[4] Department of Urology, University of Heidelberg Medical Center, Heidelberg, Germany

**Abstract.** Semantic segmentation constitutes an integral part of medical image analyses for which breakthroughs in the field of deep learning were of high relevance. The large number of trainable parameters of deep neural networks however renders them inherently data hungry, a characteristic that heavily challenges the medical imaging community. Though interestingly, with the de facto standard training of fully convolutional networks (FCNs) for semantic segmentation being agnostic towards the 'structure' of the predicted label maps, valuable complementary information about the global quality of the segmentation lies idle. In order to tap into this potential, we propose utilizing an adversarial network which discriminates between expert and generated annotations in order to train FCNs for semantic segmentation. Because the adversary constitutes a learned parametrization of what makes a good segmentation at a global level, we hypothesize that the method holds particular advantages for segmentation tasks on complex structured, small datasets. This holds true in our experiments: We learn to segment aggressive prostate cancer utilizing MRI images of 152 patients and show that the proposed scheme is superior over the de facto standard in terms of the detection sensitivity and the dice-score for aggressive prostate cancer. The achieved relative gains are shown to be particularly pronounced in the small dataset limit.

## 1 Introduction

Datasets of the size required to train high capacity FCNs for semantic segmentation are a luxury that is barely available to the medical imaging community.

Inevitably, this leads to the question of how we can optimally leverage the information in medical datasets that are as small as they currently come.

The standard approach to training FCNs relies on a per-pixel formulation of the loss, treating individual pixels in the label maps as conditionally independent from all others, and thus squanders valuable extra information. Recent work in the field of semantic segmentation was dedicated towards introducing 'structure' to deep nets, for which mostly integrated or second-stage conditional random fields (CRFs) have been considered [1, 2]. By utilizing an additional CRF, such approaches come at the cost of computational efficiency during inference. A novel and barely explored approach however aims for lending 'structure' to deep models *during training* and employs adversarial networks [3, 4].

We hypothesize that penalizing global dependencies in the label maps during training leverages complemetary information, which the standard cross-entropy training cold-shoulders. We test this hypothesis in the context of prostate segmentation and in particular the segmentation and thus detection of aggressive prostate cancer (PC). A large body of studies for the delineation of PC from MRI has been reported in the literature to date. Most of them operate voxel-wise on pre-segmented regions of interest, e.g. the prostate as a whole. The approaches differ in the applied classification methods ([5–12]) as well as in the additional auxiliary steps they encompass (manual or thresholded prostate segmentation, feature extraction, feature selection and post-processing of the tumor segmentation, e.g. using CRFs [5, 9, 13]).

In contrast to aforementioned work, we propose a joint FCN-based segmentation of the prostate's regions along with the targeted cancer nodules that is learned in an end-to-end, thus fully automatic, fashion using *purely* adversarial training. Auxiliary steps like candidate preselection or post-processing become obsolete. As our core contribution, we demonstrate the superiority of the adversarial training scheme over the standard cross-entropy approach on the proposed segmentation task. This holds true across varied amounts of training examples and bears particularly strong relative gains in the small dataset limit.

## 2 Methods

**Adversarial Training for Semantic Segmentation** Generative adversarial networks (GANs) constitute a novel framework for estimating generative models via an adversarial process, in which a generative model $G$ and a discriminative model $D$ (e.g. both neural networks), are trained simultaneously [14]. The general idea is analogous to two models being pitted against each other, where one model counterfeits for example images and the other model estimates the probability for whether they are fake or not. This competition ideally drives both models to improve until fake images become indistinguishable from real ones. When run as a generative model, $G = G(\boldsymbol{z}) \sim p_g$ receives random noise $\boldsymbol{z} \sim p_z$ as input. After an optimal training procedure, $p_g$ can be shown to match $p_{data}$, the distribution governing the real data samples $\boldsymbol{x}$ [14]. The provably optimal training procedure is cast in form of a two-player objective:

$$\min_{G} \max_{D} \big( \mathbb{E}_{\boldsymbol{x} \sim p_{\text{data}}(\boldsymbol{x})}[\log D(\boldsymbol{x})] + \mathbb{E}_{\boldsymbol{z} \sim p_{z}(\boldsymbol{z})}[\log(1 - D(G(\boldsymbol{z})))] \big) \qquad (1)$$

GANs have proven enormously successful in generative applications such as image synthesis. Conditional GANs were introduced for solving ill-posed problems such as text-to-image translation [15], image-to-image translation [3] or single image super-resolution [16]. Conditional GANs receive, alongside $\boldsymbol{z}$, an additional non-random input to condition on.

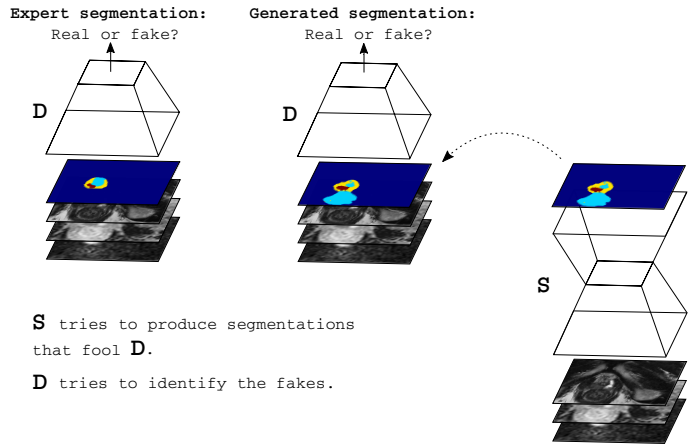

Fig. 1: Schematic illustration of adversarial training for semantic segmentation. When run deterministically $G$ is equivalent to $S$.

For this reason they are closely related to the task of semantic segmentation with $D$ being interpretable as a learned higher-order loss, a potential that has recently been realized [3,4]. The advantages of adversarial training are that it does not introduce additional complexity to the model and requires no manual design of higher-order losses, resulting in very efficient models. In this paper, we propose to use a *purely* adversarial training for FCNs. In the de facto standard training for semantic segmentation, FCNs are trained minimizing a multi-class cross-entropy loss $\mathcal{L}_{\text{mce}}$ that penalizes, for each pixel $j$ of $M$ and for each sample $i$ in a minibatch of size $N$, deviations from the correct target label vector $\boldsymbol{y}_{i,j}$:

$$\mathcal{L}_{\text{mce}}(\boldsymbol{\theta}_S) = -\frac{1}{N \cdot M} \sum_{i}^{N} \sum_{j}^{M} \boldsymbol{y}_{i,j}^{\intercal} \log S_j(\boldsymbol{x}_i), \qquad (2)$$

where we refer to what was the generator network $G$ before as segmentor $S$ in order to acknowledge the non-generative nature of our approach. The adversarial training scheme by contrast requires a discriminator $D$ that is trained alongside the segmentor $S$. In line with Eq. 1, $D$'s loss can be formulated as follows:

$$\mathcal{L}_D(\boldsymbol{\theta}_D, \boldsymbol{\theta}_S) = -\frac{1}{N} \sum_i^N \Big( \log D(\boldsymbol{y}_i) + \log\left(1 - D(S(\boldsymbol{x}_i))\right) \Big) \tag{3}$$

According to Eq. 1, $S$ then minimizes $\mathcal{L}_S = -\mathcal{L}_D$. We however follow [14] and use the loss-term below for the sake of larger gradient signals in the case when the adversary $D$ very accurately classifies real and fake segmentations:

$$\mathcal{L}_S(\boldsymbol{\theta}_D, \boldsymbol{\theta}_S) = -\frac{1}{N} \sum_i^N \log D(S(\boldsymbol{x}_i)) \tag{4}$$

This scheme is visualized in Fig. 1. Luc *et al.* [4] propose a hybrid loss term for the segmentor $S$ in form of a weighted sum, $\mathcal{L}'_S(\boldsymbol{\theta}_D, \boldsymbol{\theta}_S) = \mathcal{L}_S + \lambda \mathcal{L}_{mce}$, which we also compare against below. Optimal training requires for $D$ to be near its optimal solution at all times. For this purpose, $D$ can be trained using $k$ minibatch gradient descent steps for each such step performed on $S$ [14].

**MRI dataset** The employed dataset contains 152 patients with MRI acquired using a Siemens Prisma 3.0 T machine at the National Center for Tumor Diseases (NCT) in Heidelberg, Germany. All patients had a suspicious screening result and a core biopsy yielding pathological classification, i.e. Gleason Score (GS) [17]. Image analysis was based on a T2-weighted Image (T2w), an Apparent Diffusion Coefficient (ADC) map and a high b-value diffusion weighted image (b1500) at $b = 1500\,\text{s}\,\text{mm}^{-2}$. The T2w images have an in-plane resolution of 0.25 mm, the other two modalities were upsampled accordingly. The prostate's anatomical details as well as lesions were segmented independently on both the T2w and the ADC-map by an experienced radiologist. The annotations comprise – if present – three classes: tumor lesion, peripheral zone and transitional zone. Registration was performed using rigid translation maximizing the overlap between the PZ masks. The two independent segmentations were fused by label consensus.

**Training** To provide meaningful comparison, the training protocol is the same for all evaluated schemes. We use a set of 55 patients ($\mathcal{S}_{agg}$) comprising 188 2D-slices with biopsy-confirmed aggressive tumor lesions of GS $\geq 7$ and 97 patients ($\mathcal{S}_{free}$) with 475 2D-slices that were diagnosed lesion free (slice size $3\times416\times416$). The experiments are performed using four-fold cross-validation on $\mathcal{S}_{agg}$ with mutually exclusive subject allocation to the folds, while $\mathcal{S}_{free}$ is used during training only. In each cross-validation permutation, 2 folds are employed for training the model, one fold for model selection according to the tumor dice, and one held-out fold for validation. All segmentation models are trained for 225 epochs, with 80 randomly sampled batches each, using an initial learning rate (LR) of $10^{-5}$, that is halved every 75 epochs. During the adversarial training scheme we train the discriminator $D$ on 3 batches for each batch the segmentor is trained on while using fixed LR $= 10^{-5}$ for $D$. For parameter optimizations we use *Adam* [18]. The training data is augmented by in-plane rotations with angle $\phi \sim \mathcal{U}\left[-\pi/8, \pi/8\right]$, crops with a mask shifted by $(\Delta x, \Delta y) \sim (\mathcal{U}\left[-50, 50\right], \mathcal{U}\left[-50, 50\right])$ and random left-right mirroring. We use a batch-size of 5 with importance sampling, averaging to 3.5 samples from $\mathcal{S}_{agg}$ in each batch.

**Network Architectures** We use an identical 'U-Net'-type architecture for the segmentor in each experiment [19]. We follow [3] and use InstanceNorm instead of BatchNorm, conjecturing that it avoids harmful stochasticity, introduced by small batch-sizes. Let `CLk` denote a Convolution-InstanceNorm-leakyReLU layer with $k$ filters and `Ck` denote a Convolution-InstanceNorm-ReLU layer. Then the segmentor's encoder takes on the following form: `CL64-CL128-CL256-CL512-CL1024`, while the decoder can be represented as: `C512-C256-C128-64-C4`. The architecture used for the discriminator in large parts mirrors that of the segmentor's encoder: `CL64-CL128-CL256-CL512-CL512-CL1024-GPD1`, where `GPD1` denotes a global average pooling layer followed by a dense layer with one output node. InstanceNorm is neither applied to the first nor the last layer in $S$ and $D$. Convolutional layers employ 3×3-filters, except for the last one in $S$'s decoder which uses 1×1-filters. $D$ takes 7×416×416 inputs, featuring three channels for the MRI modalities and four channels encoding the class labels.

## 3  Results

The adversarial approach scored significantly better for tumor segmentation both in the Dice coefficient (DSC) as well as the sensitivity (Tab. 1, $p < 0.001$ using Wilcoxon signed-rank test). The specificites between the approaches were equal. Fig. 2 illustrates examplary segmentations. Using a hybrid loss with the same weighting as [4] does not provide further improvements. In order to evaluate how the training schemes compare on progressively smaller datasets, we successively take away positive training samples from the fold that both schemes coincided to perform best on. We train in the exact same manner as described above and evaluate on the same held-out fold from before. The results are depicted in Fig. 3.

Table 1: Experimental results of the four-fold cross-validation for GS $\geq$ 7 Tumor.

| training scheme<br>loss | cross-entropy<br>$\mathcal{L}_{mce}$ | adversarial<br>$\mathcal{L}_S$ & $\mathcal{L}_D$ | hybrid<br>$\mathcal{L}_{mce}/2 + \mathcal{L}_S$ & $\mathcal{L}_D$ |
|---|---|---|---|
| tumor DSC | $0.35 \pm 0.29$ | $\mathbf{0.41 \pm 0.28}$ | $0.39 \pm 0.29$ |
| tumor sensitivity | $0.37 \pm 0.33$ | $\mathbf{0.55 \pm 0.36}$ | $0.49 \pm 0.35$ |
| tumor specificity | $0.98 \pm 0.14$ | $0.98 \pm 0.14$ | $0.98 \pm 0.14$ |

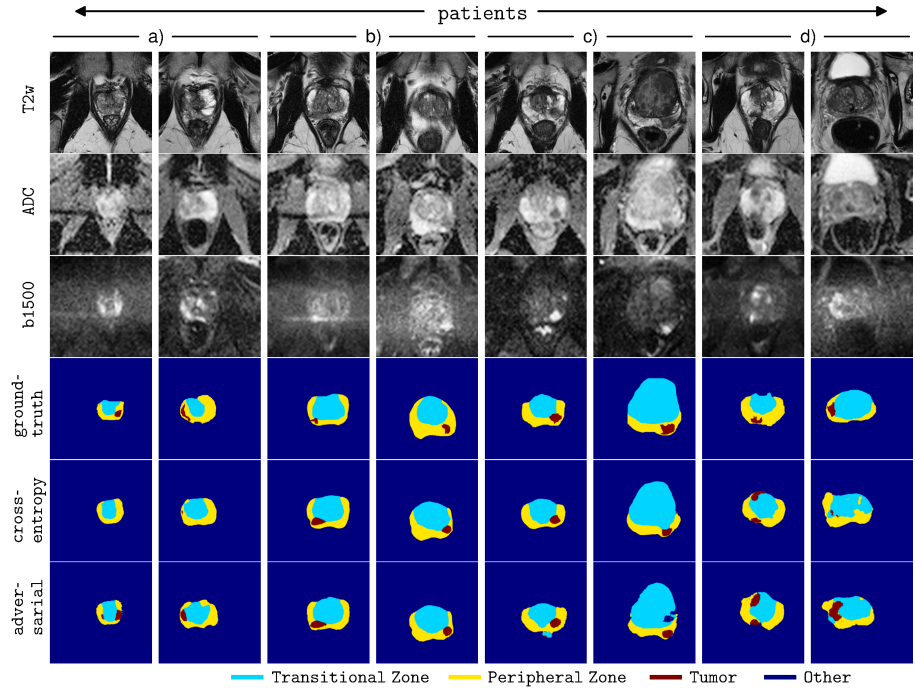

Fig. 2: Examples depicting the three MRI modalities, the expert annotation as well as the segmentations produced by training the segmentor network with different loss schemes. The first two columns from the left, i.e. columns a), depict examples in which the adversarial is clearly more sensitive to aggressive tumor than the cross-entropy training. Columns b) show examples for which the methods are on par. Columns c) feature examples for which the adversarial method yields partially defective label maps. Columns d) exhibit examples for which both methods deviate considerably from the ground-truth, the first of which likely shows tumor detection by both methods, missed by the expert.

## 4 Discussion

To the best of our knowledge we are the first to introduce the concept of adversarial training for semantic segmentation of medical images. The adversary $D$ constitutes a learned parametrization capturing the essence of what amounts to a plausible segmentation – information that is not harnessed in the conventional cross-entropy training. Our experiments show that the proposed method is more efficient and increases detection sensitivity and the dice-score of aggressive prostate cancer, a segmentation task that is challenging due to the strong tissue heterogeneities of the prostate and the subtle tumor appearance.

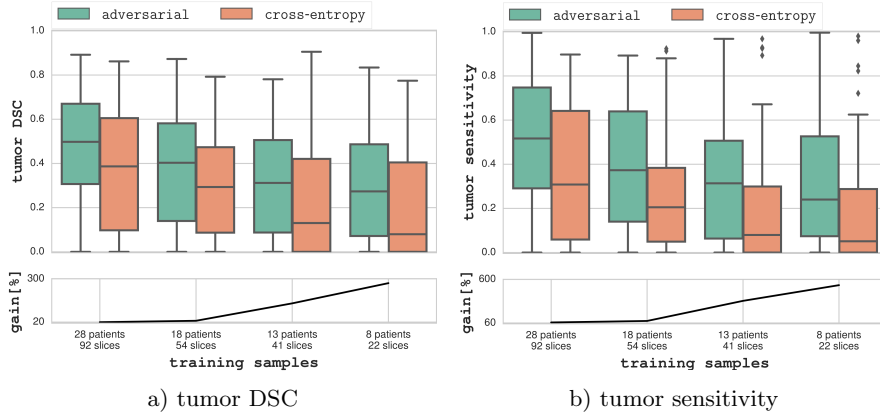

Fig. 3: Comparison of the performance measured in terms of tumor DSC (a) and sensitivity (b) between the adversarial and cross-entropy training when successively taking away training data. The upper panels illustrate the respective distributions for the two schemes. The lower panels show the relative gain in median of the adversarial over the cross-entropy training, from which particularly pronounced gains are visible in the small dataset limit. Specificity (not shown) was around 0.98 in all experiments.

The novelty of the presented approach also lies in the precise formulation of the loss, which is, in contrast to the pioneering publications [3, 4], *purely* adversarial. Our work further differs in the architectural details of the models, e.g. we employ concatenation of the MRI images and the labels for direct spatial correspondence in the input of the adversary.

Previous approaches for prostate cancer detection propose a significantly more involved series of steps including pre-determination of regions of interest and post-processing. Because these methods are almost always evaluated in a pre-delineated region and the study populations are considerably smaller than ours, comparison is severely limited. An exception being [10], who use a dataset of 347 patients, but perform evaluations on a tumor candidate level. For completion, our DSC result of $0.41 \pm 0.28$ may be compared against the SVM+CRF based approaches of [5] and [13] who report a DSC of $0.46 \pm 0.26$ (which we reach on individual folds) and 0.39, but only use 23 and 20 patients respectively.

The limitations of our work include possible mismatches between pathology and expert annotation due to several reasons including registration errors and ob-server variability. Furthermore, bleeding-edge architectures and an extension of the training scheme to 3D models remain to be explored. Our contribution could inspire future developments in and around the detection of aggressive PC in particular as well as in semantic segmentation of medical images in general.